# Pollen13K: A Large Scale Microscope Pollen Grain Image Dataset

**Sebastiano Battiato**
University of Catania
Viale A. Doria, 6 Catania – 95125
sebastiano.battiato@unict.it

**Alessandro Ortis**
University of Catania
Viale A. Doria, 6 Catania – 95125
ortis@unict.it

**Francesca Trenta**
University of Catania
Viale A. Doria, 6 Catania – 95125
francesca.trenta@unict.it

**Lorenzo Ascari**
University of Turin, Italy
lorenzo.ascari@unito.it

**Mara Politi**
University of Turin, Italy
mara.politi@unito.it

**Consolata Siniscalco**
University of Turin, Italy
consolata.siniscalco@unito.it

## Abstract

Pollen grain classification has a remarkable role in many fields from medicine to biology and agronomy. Indeed, automatic pollen grain classification is an important task for all related applications and areas. This work presents the first large-scale pollen grain image dataset, including more than 13 thousands objects. After an introduction to the problem of pollen grain classification and its motivations, the paper focuses on the employed data acquisition steps, which include aerobiological sampling, microscope image acquisition, object detection, segmentation and labelling. Furthermore, a baseline experimental assessment for the task of pollen classification on the built dataset, together with discussion on the achieved results, is presented.

## 1 Introduction and Motivations

Aerobiology, the discipline that studies airborne biological particles and their dispersal mechanisms, has a crucial role in several fields such as medicine, biology, and agronomy, with direct and non-direct effects on the economy and public health. Estimating the abundance of airborne allergenic pollen and fungal spores allows to evaluate the associated health risk and the potential infectious diseases [1] on both humans [2] and plants [3] for certain periods. The amount of airborne pollen can be considered as a proxy to plant phenology and flowering intensity, thus leading to its integration in many yield forecasting systems applied to commercially important crops [4, 5]. Despite its effectiveness, the involvment of an expert to analyze images in microscopy is a time-consuming task that has hindered the application of aerobiology to those and new sectors [6]. Despite of the various efforts to develop devices that allow the identification and classification of pollen grains without the need of end-user intervention [7, 8], the observation and discrimination of features from relevant entities performed by qualified experts is still predominant [9]. Recently, several tools have been developed to accomplish tasks related to the identification and the classification of pollen grains by using the modern Deep Learning techniques. Indeed, Deep Learning has produced impressive results in different fields [10, 11, 12], taking advantage of a large amount of labeled data. Specifically, the spread of the methods based on deep neural networks has led to the definition of large-scale datasets which are useful to obtain more reliable results. To this aim, we have constructed a dataset of more than 13.000 objects from microscope pollen grain images.

Table 1: Comparison between the proposed dataset and the main datasets used in pollen grain classification.

| Dataset | Number of Grains | Image Type | Resolution |
|---|---|---|---|
| **Duller's Pollen Dataset [13]** | 630 | Grayscale | 25x25 |
| **POLEN23E [14]** | 805 | Color | Minimum 250 pixel per dimension |
| **Ranzato et al. [15]** | 3.686 (1.429 images) | Color | 1024x1024 (multiple grains per image) |
| **Proposed Dataset** | >12.000 + ~1.000 examples of debris (e.g., dust, air bubbles) | Color | 84x84 |

Table 2: Number of objects for each class.

| Class | Label | Number of Objects |
|---|---|---|
| *Corylus avellana* (well-developed pollen grains) | 1 | 1.850 |
| *Corylus avellana* (anomalous pollen grains) | 2 | 903 |
| *Alnus* (well-developed pollen grains) | 3 | 9.558 |
| Debris | 4 | 999 |
| Cupressaceae | 5 | 43 |

Specifically, we developed a proper pipeline to detect and extract four classes of the pollen grain and an additional class of objects, called Debris, which includes air bubbles, dust, etc. We applied the image processing pipeline to digitalized microscope images acquired from aerobiological samples. The built dataset is available for research purposes [1]. In this instance, the aim of the present study is to benefit the future research into pollen classification providing an effective pipeline for assisting experts at classifying object into several species. The remainder of this paper is organized as follows. Section 2 lists existing work in which public dataset have been used for automatic pollen grain classification. In addition, we described the composition of the proposed dataset. In Section 3, we described the segmentation pipeline. Section 4 describes how we applied a pool of standard Machine Learning approaches to perform automatic pollen grains classification focusing on preliminary results. In Section 5, we reported the final considerations of this paper.

## 2 Dataset

Previous studies have investigated the problem of automatic pollen grain detection and classification in which selfcollected datasets have been used to evaluate the proposed pipelines. Two public databases are the Duller's Pollen Dataset [13] and the POLEN23E [14]. The first, contains a total of 630 grayscale images of size 25x25, the latter includes 805 color images of 23 pollen species, with 35 images for each pollen type. Larger datasets have been proposed for the task of pollen grain detection such as the one presented in [15]; however, the number of grains is just 3,686. All related works are summarized in Table 1. In our study, a total of 13,416 objects from aerobiological images were segmented under the guidance of experts. They collected the airbone samples which were processed and analyzed using a longitudinal read of adhesive tapes. Each tape was set on a rotating drum, moved at 2 mm h-1 under a suction hole, and the pollen grains that adhered to it were inspected on a daily basis segments by using a Leitz Diaplan brightfield microscope and a 5 MP CMOS sensor. In accordance with the procedures standardized for this process, the pollen walls placed on the microscope slides were selectively stain with a mounting medium containing basic fuchsin (0.08 % gelatin, 0.44% glycerin, 0.015% liquefied phenol, 0.0015% basic fuchsin in aqueous solution). Acquired images were affected by a heavy background noise deriving either from the aerobiological sample itself (i.e. debris and dust and fungal spores) or from the mounting technique (air bubbles). For this reason, we developed a segmentation pipeline to locate and extract objects from the microscope images. We collected a RGB image for each extracted object (84x84 resolution), alongside its binary mask and segmented image with green background. Experts in aerobiology field manually labelled segmented objects grouping them into five different categories: *Corylus avellana* (well-developed pollen grains), *Corylus avellana* (anomalous pollen grains), *Alnus* (well-developed pollen grains), Cupressaceae, Debris (bubbles, dust and any non-pollen object). In Fig. 1, examples of objects for each class are reported. Under the supervision of a team of experts in aerobiology, we discarded 63 objects. Specifically, images depicting pollen objects overlapped to non-pollen ones (i.e., bubbles), which could lead to ambiguous labelling. Hence, the resulting number of segmented objects is 13,353. Furthermore, the number of objects per class does not make up an equal portion of dataset. In Table 2, we reported the amount of detected objects per class.

---

[1] Dataset website: https://iplab.dmi.unict.it/pollengraindataset/

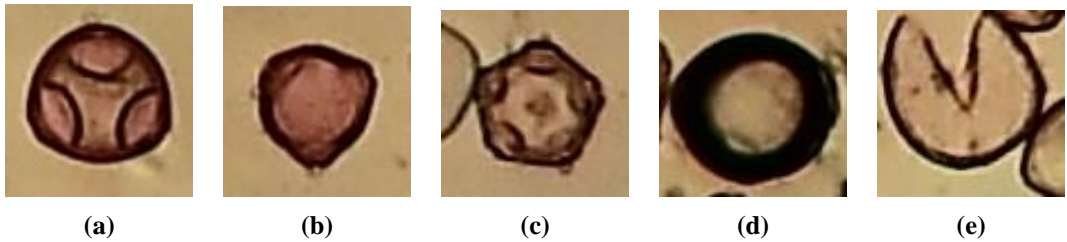

Figure 1: Examples of acquired samples. (a) *Corylus avellana* (well-developed pollen grains), (b) *Corylus avellana* (anomalous pollen grains), (c) *Alnus* (well-developed pollen grains), (d) Debris, (e) Cupressaceae.

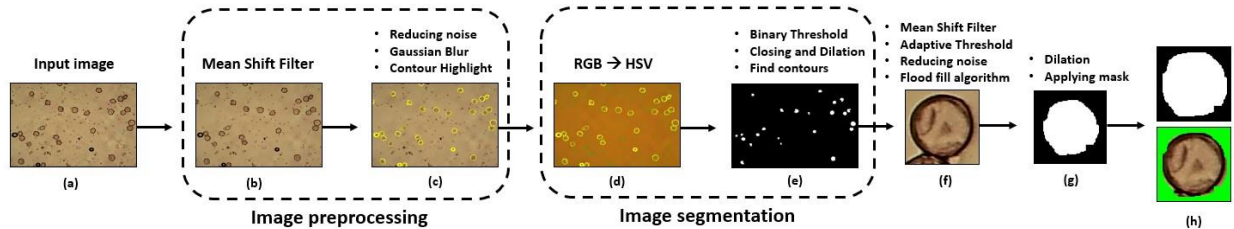

Figure 2: The overall pipeline. (a) Image of an aerobiological sample, (b) Obtained image after applying mean shift filtering function, (c) output image after reducing background noise and smoothing it with a Gaussian filter, (d) the resulting image after converting color space from RGB to HSV, (e) the mask generated by applying binary threshold, closing and dilate operators, (f) the segmented object after detecting object contours from previous mask image, (g) the resulting binary mask after applying a mean shift filter and adaptive threshold, (h) the obtained binary image and the related segmented object with green background.

## 3  Segmentation pipeline

The proposed pipeline can be split into two main blocks: (i) image pre-processing, (ii) object segmentation using morphological operations. The overall scheme of the proposed method is depicted in Fig. 2.

### 3.1  Pre-processing Pipeline

The aim of the pre-processing pipeline is to improve the quality of input images in order to highlight objects contours by reducing also the amount of background noise present in aerobiological images. In the first stage, we applied a mean shift filtering to provide an image with a flat coloured texture. For a given pixel *(x,y)*, the mean shift algorithm finds the potential neighbor *(X, Y)* not only taking into account its spatial position but also its color hyperspace.

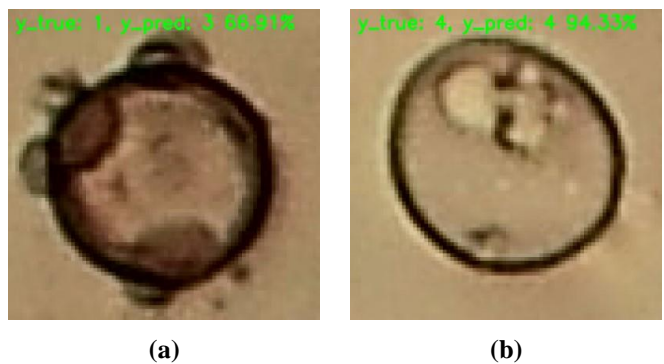

Figure 3: Examples of segmented object classification. (a) Mis-classification of a well-developed pollen object. (b) Correct classification of a Debris object.

This step is crucial for all the following steps having implications in the performance of image segmentation. The second block is designed to smooth background in order to maximize the object detection. After converting the input image to a grey level image, we applied binary thresholding combined with Otsu's method [16] (threshold value set to 127). Since pollen objects were usually higher than 500 pixel, we removed all the objects which were less than this size analyzing connected components with 8 neighbors. Then, we applied the resulting binary mask to the input image and colored the detect contours in yellow. The key point of creating colored contours is to distinguish which objects are inside the foreground area and which are not. Finally, a $11 \times 11$ kernel Gaussian filter is applied to blur the input image and reduce details in background. We combined the obtained images from both previous steps with the aim of highlighting objects placed on the foreground against the ones in background.

## 3.2 Segmentation Pipeline

Objects segmentation is the second and important step of the overall pipeline. Firstly, we converted the output image of previous stage from RGB color model to HSV color space. Then, we performed binary thresholding using grayscale input image. In order to reduce the noise background generated by previous image processing steps, we applied a closing operator followed by a dilation using the same $3 \times 3$ kernel for both operations. Then, we used flood fill algorithm to reassign values of all neighbouring pixels of a given point with a required uniform color. In this instance, we implemented flood fill algorithm with the aim of distinguishing the foreground from the background. Hence, all the objects of interest have been filled with a black color whereas the background has been filled with a white color. After removing objects with a size smaller than 100 pixel by analyzing connected components in image, we found objects contours by using the obtained binary mask. Furthermore, the coordinates *(X, Y)* of their centers were calculated. We observed that the obtained binary masks presented holes or were considerably smaller than the region of the object of interest. In order to improve the overall quality of binary images, we applied a mean shift filtering function and converted the input image in grey mode. Hence, we used an adaptive threshold which takes an adaptive threshold Gaussian as input parameter. The Gaussian threshold value is a Gaussian-weighted sum of the neighbourhood values minus the constant C. Based on this, we set the block size (neighbourhood) parameter to 77 and C value to 0. At this stage, we reduced noise through the use of connected components for selecting objects with a size greater than 150 pixel and we applied flood fill algorithm . Finally, we used a dilation operator selecting a $3 \times 3$ kernel with full of ones and a number of iterations equal to 5 to increase the size of object in binary mask image. Segmentation results, in terms of number of detected objects and clearness of segmentation masks, have been evaluated by experts.

# 4 Preliminary classification results

## 4.1 Experiments using Machine Learning classifiers

In our study, we first considered a pool of Machine Learning methods with the aim of performing an effective pollen grains classification. To further investigate textures, Local Binary Pattern (LBP) [17] and Histogram of Oriented Gradient (HOG) [18] were computed to generate feature vectors from images. In order to conduct our experiments, we used the set of segmented images with green background. We subdivided this dataset by selecting 85% of images as training set and 15% of images as test set. Also, we carried out our experiments by using the following models: Linear Support Vector Machine (SVM), RBF SVM, Multi-Layer Perceptron (MLP), Random Forest, AdaBoost [19]. For each model, we identified optimal parameters by using Grid Search algorithm performing 10 trials and computing the average accuracy obtained at each run. Finally, we assessed the effectiveness of each classifier using our imbalanced dataset and relying on optimal parameters that have been selected from the previous stage. In class imbalanced classification, the major issue involves the lack of samples of a given class which produces inaccurate results. In this instance, the standard performing metrics, such as accuracy, could become an unreliable measure of model performance. To furher address the imbalance in the data, we considered penalized classification models for SVM and Random Forest, which adjust weights inversely proportional to class frequencies in the training data. These penalties help the model to pay more attention to the minority class. For our experiments, we implemented a stratified train-test splitting to ensure that such classes are equally balanced in both training and test set. Moreover, considering the small number of observations related to Cupressaceae class (43), we did not include them in the dataset used for the experiments. To evaluate the performance of each classifier, we used the weighted F1 score for quantitative evaluation [20]. This metric has been selected considering the imbalance in the data. The weighted F1 score function calculates the F1 metrics for each class, and their average weighted by support (the number of true instances for each class). With regard to HOG, our experiments have shown that the best classification results were obtained by using a RBF SVM [21]. In Table 3, we reported the best results in terms of accuracy and F1 score for each classifier. We observed that RBF SVM achieved an accuracy and a F1 score over 85% with a gamma value of 0.1 and a C value of 1000. Furthermore, the MLP model consistently yielded to a F1 score of 0.8431 with alpha value equal to 0.1 and a number of estimators equal to 300. Nevertheless, we observed that SVM with RBF kernel did not have a significant classification performance when using LBP features. Specifically, the lowest F1 score is equal to 67% considering a gamma value of 1.0 and a C value of 1000.

July 9, 2020

Table 3: Comparison between the best results by using HOG and LBP features.

| Methods | Parameters | | HOG | |
|---|---|---|---|---|
| | | | Accuracy | F1 score |
| **LINEAR SVM** | C = 1000 | | 0.7646 | 0.7673 |
| **RBF SVM** | G = 0.1 | C = 1000 | **0.8658** | **0.8566** |
| **RANDOM FOREST** | EST = 10 | | 0.7616 | 0.7124 |
| **ADABOOST** | LR = 0.5 | EST = 500 | 0.7752 | 0.7627 |
| **MLP** | a = 0.1 | EST = 300 | 0.8493 | 0.8431 |
| **Methods** | **Parameters** | | **LBP** | |
| | | | Accuracy | F1 score |
| **LINEAR SVM** | C = 100 | | 0.7446 | 0.7439 |
| **RBF SVM** | G = 0.1 | C = 1000 | 0.6430 | 0.6714 |
| **RANDOM FOREST** | EST = 1000 | | 0.7792 | 0.7387 |
| **ADABOOST** | LR = 1.0 | EST = 100 | 0.7722 | 0.7487 |
| **MLP** | a = 0.0001 | EST = 500 | 0.8002 | 0.7764 |

### 4.2 Experiments using Deep Learning models

After evaluating the results of previous experiments, we performed object classification through the use of a Deep Convolutional Neural Network (i.e., AlexNet [22]). In this instance, we carried out our experiments using two different settings for the training data. To create the first set, we only included objects with noisy background whereas for the second one, we defined an augmented dataset inserting both sets of images with green and noisy background. Subsequently, AlexNet was trained on these new datasets. For both experiments, we also performed data augmentation in training data. Moreover, we set the base learning rate to 0.001, the batch size to 64 and number of epochs to 1000. We also introduced an Early stopping function to stop training once the model performance stops improving in order to manage overfitting problem. Then, we evaluated the network performance on the test set after every 10 epochs. AlexNet achieved an average F1 score of 0.87 using the auugmented dataset. With regard to the first dataset, it could be observed that AlexNet achieved an average F1 score of 0.74. In Fig. 3, we illustrated two examples of classification performed by AlexNet. Our experiments showed that AlexNet tends to mis-classify pollen objects of *Corylus Avellana* (well-developed pollen) class which present a texture similar to object of *Alnus* class. With regard to Debris object, AlexNet is able to classify it accurately. For the sake of completeness, we performed our experiments by using also a different CNN architecture which could represent a more appropriate solution for analyzing images with a small size. In our case, we implemented a SmallerVGGNet which is a more compact variant of Very Deep Convolutional Networks (VGGNet) [23]. For these experiments, we set the same parameters used for AlexNet. SmallerVGGNet achieved an average F1 score of 0.85 using augmented dataset and an average score of 0.69 using images with noisy background [24].

## 5 Conclusions

In this work, we presented a large-scale pollen image dataset composed of more than 13,000 objects. In particular, we described the main stages to detect and extract the objects from a set of microscope images. The presented project, based on automatic pollen grain classification, has involved experts in aerobiology and Computer Vision field. More specifically, aerobiologists have manually labeled the segmented objects by grouping them into five different categories. Then, we performed the pollen grain classification by using the most common Machine Learning techniques (e.g., SVM, AdaBoost, etc.) in order to evaluate their performances. In our future works, we are planning not only to improve the image segmentation pipeline but also define an effective Deep Learning architecture to improve the classification results.

## 6 Acknowledgements

The research has been carried out thanks to the collaboration with Ferrero HCo, that financed the project and allowed the collection of aerobiological samples from hazelnut plantations.